\documentclass{article}
\begin{document}

\title{Why should we ever automate moral decision making?\thanks{This short writeup accompanies an invited talk at the workshop on {\em Ethics and Trust in Human-AI Collaboration: Socio-Technical Approaches}, August 21st, 2023, Macao.  That talk was titled {\em Automated Moral Decision Making by Learning from Humans: Why and How.}}}
\author{Vincent Conitzer\\Carnegie Mellon University \& University of Oxford}
\date{}

\maketitle

\begin{abstract}
While people generally trust AI to make decisions in various aspects
of their lives, concerns arise when AI is involved in decisions with
significant moral implications. The absence of a precise mathematical
framework for moral reasoning intensifies these concerns, as ethics
often defies simplistic mathematical models. Unlike fields such as
logical reasoning, reasoning under uncertainty, and strategic
decision-making, which have well-defined mathematical frameworks,
moral reasoning lacks a broadly accepted framework. This absence
raises questions about the confidence we can place in AI's moral
decision-making capabilities.

The environments in which AI systems are typically trained today seem
insufficiently rich for such a system to learn ethics from scratch,
and even if we had an appropriate environment, it is unclear how we
might bring about such learning. An alternative approach involves AI
learning from human moral decisions. This learning process can involve
aggregating curated human judgments or demonstrations in specific
domains, or leveraging a foundation model fed with a wide range of
data. Still, concerns persist, given the imperfections in human moral
decision making.

Given this, why should we ever automate moral decision making -- is it
not better to leave all moral decision making to humans? This paper
lays out a number of reasons why we should expect AI systems to engage
in decisions with a moral component, with brief discussions of the
associated risks.
\end{abstract}

\section{Introduction}

People are generally quite comfortable with AI making all kinds of
decisions in their lives.  We are happy for AI to choose a route for
us to follow when driving, to choose which articles we read or which
videos we see, or even to propose people for us to date.  But we often
feel less comfortable about the use of AI in settings where there is a
significant moral component to the decision.

One good reason to be concerned about this is that we do not currently
have a clean, mathematically precise framework for moral reasoning.
Indeed, much of the field of ethics concerns how simplistic
mathematical frameworks fall short.  For example, simplistic versions
of act utilitarianism might have us kill a patient with a minor
illness to redistribute that patient's organs to other patients, who
would die without those transplants.  In contexts other than ethics,
we do have clean mathematical frameworks.  For example, we have such a
framework for logical reasoning; and thanks to it, AI techniques (say,
using SAT solvers) can help us prove certain kinds of theorems (for a
recent example, see~\cite{Subercaseaux23:Packing}).  Similarly, we have
such a framework for reasoning under uncertainty (the theory of
probability, graphical models, probabilistic programming, etc.); and thanks
to it, we have applications such as monitoring the Comprehensive
Nuclear-Test-Ban Treaty~\cite{Arora11:Global}.  And we have
such a framework for strategic reasoning (game theory); and thanks to
it, we now have, for example, superhuman-level poker
AI~\cite{Brown19:Superhuman}.  But, again, we lack such a framework
for {\em moral} reasoning -- or, to the extent we have such
frameworks, they are highly controversial and not broadly endorsed.
And it seems unlikely that we will find such a framework
soon.\footnote{Perhaps mathematical frameworks that narrowly focus on
  one particular ethical issue -- e.g.,~\cite{Letchford08:Ethical} --
  are more likely to be successful, but these will necessarily have
  limited use as well.}

Might we be confident in the quality of AI's moral decision making
without such a mathematical framework?  It is conceivable, in
principle, that in a sufficiently rich environment, AI could learn
ethics from scratch.  But it seems unlikely that any environments in
which AI systems are trained today are sufficiently rich; perhaps
something like Melting Pot~\cite{Leibo21:Scalable} has
many relevant aspects, but still is likely to fall short.  Moreover,
even if we did have a sufficiently rich environment, it is not clear
that we currently know how to train AI systems in such an environment
in a way that lets them learn ethics from scratch.

For now, perhaps the most promising approach is for AI systems to
learn moral decision making from human beings~\cite{Conitzer17:Moral}.
They could learn that by aggregating curated human judgments or
demonstrations in specific
domains~\cite{Noothigattu17:Making,Kahng19:Statistical,Freedman20:Adapting},
or perhaps from a very broad set of data through a foundation
model~\cite{Hendrycks21:What}.  Still, it is natural to have concerns
about this approach, especially given that human moral decision making
is surely not perfect.

\section{Reasons for automated moral decision making}

Given the above, one may well wonder why we should be interested in
automated moral decision making at all; can we not simply leave all
moral decision making to human beings?  In what remains, we cover some
reasons why, in spite of the lack of a formal framework, we may yet
want to have AI systems do automated moral reasoning, rather than
simply leaving the relevant decisions to a human being.  As we will
discuss, some of these reasons overlap, and some uses of AI can call
on multiple of these reasons for support.  So, these reasons are not
intended to be disjoint from each other; and presumably this list does
not exhaust all such reasons.  For example applications given below, I
will not try to argue that the benefits of automated moral reasoning
outweigh the downsides.  The intent is that the list of reasons below
would be useful even to someone who is opposed to the deployment of AI
in any of these applications, if only to understand why others might
nevertheless choose to proceed with such deployment.

While each possible application of automated moral decision making
comes with its own risks, the reason why automated moral decision
making is used in the first place is often informative of the risks
faced and the ways to mitigate those risks.  Therefore, along with
each reason, we present a brief analysis of the risks associated with
using automated moral decision making for that reason.  There may of
course be risks associated with humans making these decisions as well,
but we will not get into those here.  The first three of the following
reasons were also briefly discussed in~\cite{Freedman20:Adapting}.

\subsection{Speed}

In some cases, decisions need to be made {\em faster} than humans can
make them, or faster than humans can make them well.  One example is a
self-driving car that suddenly faces an unexpected situation.  For
instance, an accident occurs immediately in front of it, and it needs
to make a decision to either brake hard at some risk to the occupants,
or attempt to swerve around the accident, which runs the risk of
colliding with an adjacent car (possibly depending on the reaction of
the adjacent car).  This a moral dilemma.  But passing control back to
the human occupant, who does not have situational awareness, is likely
to do little good.  Cybersecurity and cyberwarfare provide additional
examples where speed can be of the essence.  One might argue that this
is a reason to avoid these types of scenarios altogether -- maybe we
should not have self-driving cars at all, and maybe we should work
harder to ensure our systems are not vulnerable to cyberattacks or
otherwise prevent such attacks from happening, etc.  This is not the
place to get into these discussions; all we aim to argue here is that
these are settings where simply having a human take over the moral
reasoning at the moment that it is needed is not likely to address the
problem.

{\bf Risks.}  When AI is adopted for the sake of being able to act
faster, there does not seem to be any inherent limit on how bad the
consequences can be, because there will be no chance for a human to
review the decision.

\subsection{Scale}

In some settings, {\em many} decisions need to be made, and it simply is
not reasonable to have a human make each of these decisions.  For
example, consider the decision of whether to show someone a
potentially sensitive ad, where the ideal decision requires taking
into account detailed features of the user.  This, too, can be a moral
dilemma.  The impact of any one such decision is likely to be small,
but as the decisions are made across millions of users, their impact
adds up.

{\bf Risks.}  In this context, it seems sensible to periodically review
a sample of the decisions; and the more important each individual
decision, the more often it makes sense to review.  If this is all done
well, it naturally limits how badly things may go.

\subsection{Complex optimization}

In some cases, moral reasoning must be intertwined with complex
optimization.  A good example of this is the problem faced in a {\em
  kidney exchange}~\cite{Roth04:Kidney}.  In kidney exchanges, AI is
already used to determine which potential kidney donors to match with
patients~\cite{Abraham07:Clearing}.  Even the problem of maximizing
the number of transplants is computationally hard, but it is not clear
that that is the correct objective to pursue; in making these decisions, maybe we
should also take into account other aspects, such as the patients'
age, other aspects of their health, perhaps even whether they have
dependents or a criminal record, etc.  Trading off these varying
aspects between feasible solutions poses a moral dilemma.

In this context, the moral reasoning problem does not seem cleanly
separable from the optimization problem in such a way that humans can
do the moral reasoning when it is needed.  In principle, the AI system
may propose a selection of feasible solutions, from which (say) a
committee of humans then picks one.  But without the AI system doing
any moral reasoning of its own, it is not clear how to guarantee that
its selection will include the best solution, or even a good one;
except, perhaps, if the selection includes {\em all} reasonable
solutions -- but there will generally be exponentially many of these,
and searching through exponentially many options is what we brought
the AI in to do for us in the first place.
(In~\cite{Freedman20:Adapting}, we propose a method for having the AI
learn an objective function from human feedback, but there remain a
variety of questions about how best to elicit such information from
human subjects~\cite{Chan20:Artificial,McElfresh21:Indecision}.)

{\bf Risks.} Here, too, the risks of automated moral decision making
could be limited by the fact that decisions can be reviewed.  One
could, for example, compare the AI-generated decision to a
human-generated one in each instance, to make sure the AI-generated
one is in fact better.

\subsection{Better world models}

In some domains, we may not have good intuitions about the actual
consequences of decisions. Consider, for example, an AI system tasked
with the design of new drugs to treat a disease.  It may have better
``intuitions'' about the effects of various potential drugs than we
do; and, for the system, choosing which new drug to propose to us
requires trading off that drug's expected efficacy with its expected
side effects.  Presumably, we will first still want to conduct a
randomized controlled trial on the drug, but even just deciding to
start such a trial is a decision with significant consequences.  We
may want to leave the decision in our own hands, and ask the AI system
to tell us the reasons for its choice of proposed drug.  But then
again, it may not be able to effectively explain these reasons to us,
for example because its model of how these drugs work does not
translate well to natural language.  Even if we as humans {\em were}
able to evaluate the pros and cons of any proposed drug, there is
still the issue of how it selects a set of proposed drugs for us to
select from; at this point we are back at the issue of complex
optimization discussed previously.

A related but different example is an AI system proposing a specific
treatment for a specific patient.  In this example, we cannot first
run a trial; we simply have to decide whether to follow the proposed
treatment or not.  Again, we may not understand the reasons why the AI
system proposed this treatment; moreover, overworked physicians may
not have the time and wakefulness to thoroughly study the proposed
treatment (see also ``humans are poor decision makers under certain
circumstances'' below).  Going one step further, the AI system may
control the treatment directly.

It may seem that these examples that involve the treatment of only a
single patient do not involve much moral reasoning, as there is no
trading off between the welfare of multiple people.  Nevertheless, the
AI system may still have to trade off, for example, the pain the
patient feels against the chances of keeping the patient alive.  In
some cases, the AI system may need to decide whether to allocate
scarce medical resources to the patient even though they could help
other patients as well.  And perhaps in some cases, we would consider it
acceptable for the AI system to try out an unusual course of treatment
in part for the purpose of learning more about this course of
treatment, to be able to help future patients better.

{\bf Risks.} To the extent that humans cannot review the decisions
effectively because we do not understand the reasons behind them,
risks seem significantly higher.

\subsection{Transparency of process}

Sometimes, it is better to have a clear policy for how to make
decisions than it is to make decisions on a case-by-case basis.  In
particular, human decision making is generally not transparent; human
decision makers can be sensitive to bias, and can even be corrupt.  In
contrast, if a clear policy is set in advance, this helps to evaluate
bias and prevent corruption on individual decisions.  It can also help
to prevent other ways in which ad-hoc decision making might be
``gamed'' by interested parties.

Committing to use a specific AI system to make decisions has at least
some of the benefits of setting a clear policy.  While AI systems vary
in their transparency and interpretability, they can generally at
least be audited by testing them on a variety of inputs, and they
cannot be bribed.

Of course, there remain major problems with AI systems displaying
unfair biases, for example due to the data they were trained on; at
the very least, much more work is needed to address these problems
effectively in such systems, and perhaps in the end we will conclude
that there should always be a human in the loop in certain
applications.  The argument here is merely that in principle, there
could be an advantage to the use of AI in terms of transparency of
process; this is not to say that no further work is needed to attain
(most) of these benefits, and it is not to say that these benefits
outweigh other concerns.

The use of AI in kidney exchanges can be seen as illustrating the
transparency-of-process benefit of using AI (as well as the integrated
optimization benefit explained earlier).  A human choosing how to
match patients and donors might, whether consciously or not, let
various biases play a role in the decision; and, at least in
principle, such a person might be bribed to make this high-stakes
decision work out better for a certain patient.

Other example applications where this benefit could play a role
include the allocation of scarce medical resources more generally, or
other key resources such as housing; as well as uses in the criminal
justice system.  Of course, the use of AI in the latter context is
extremely controversial, and much work is needed to do this in a
responsible way.  But at least in principle, the
transparency-of-process advantage may yet come to be seen as important
enough to justify the use of AI even in this context.

{\bf Risks.}  Human review of decisions interacts with this reason in
a tricky way, as the possibility of humans overruling the decision
potentially takes away much of the benefit of the transparency of the
decision process.  For that reason, we may commit ourselves not to
overrule the AI system's decisions -- but this increases the risks of
these decisions.  Perhaps a balance can be struck, for example by
allowing overruling only if a supermajority of human judges concludes
that this is the right thing to do.

\subsection{Humans are poor (moral) decision makers under certain circumstances}

This reason overlaps with the ``speed'' reason above, but there are
other circumstances under which humans can be bad decision makers ``in
the moment.''  To illustrate, imagine an AI application that, when
someone is trying to send an email after a late night partying and
drinking, has the ability to analyze that email and, if the email does
not seem wise to send, to prevent it from being sent at that point in
time.  To be effective, the AI may have to engage in moral reasoning.
For example, suppose the message the user is trying to send out late
at night is: ``I don't trust that guy you were just talking with, I
recommend that you ditch him.''  The AI may have reason to believe
that the sender would probably regret sending this message in the
morning, but, in making the decision whether to temporarily block the
message, has to trade this off against the welfare of recipient in
case the sender is in fact onto something.

There are many other conditions under which we humans are poor (moral)
decision makers.  Besides cases in which our reasoning is obviously
compromised -- say, due to alcohol consumption, fatigue, illness,
etc. -- we may also display various biases, in particular when we have
a personal stake in the decision.

{\bf Risks.}  As in the case of the ``speed'' reason above, human
review may in some cases not be possible, thereby increasing the risk.
On the other hand, in some cases (such as the imagined blocked email
message above), we could let another human quickly review the decision
if it is deemed important enough, so that if this is set up well, the
risk is limited.

\subsection{Economic efficiency}

Sometimes, we may wish to deploy AI simply to reduce costs.
(This reason often overlaps with the ``scale'' reason above.)  For
example, consider the project of further automating call centers.  To
be concrete, consider a government-run healthcare hotline.  Let us
suppose that in fact, the AI still functions poorly enough that
callers would be better served if they were instantly connected to a
well-trained human being, so that the primary purpose of using AI to
handle calls is to reduce costs.  The AI may then face morally
challenging decisions when determining which callers to connect to one
of the scarce human beings answering calls, and which callers to try
to handle itself.  While this may sound like an undesirable scenario
from the perspective of a caller, the resulting cost savings can of
course be valuable; the government could in principle take the savings
and apply them towards (say) preventive healthcare education
campaigns.

{\bf Risks.}  Similarly to the ``scale'' reason above, in this context
it seems to make sense to periodically review decisions, and to do so
more frequently the more important the decisions are; if this is done
well, then risks stay limited.

\section{Conclusion}

It may seem that it is a bad idea to have AI systems make moral
decisions, or that at the very least, they should not do so unless and
until we have an appropriate, mathematically precise theory for doing
so; and that for now, we should leave moral decision making to humans.
In this short paper, we have considered a variety of reasons for why
we might nevertheless expect AI systems to end up making decisions
with a significant moral component.  This may be cause for concern,
and we have also discussed various risks associated with it; and just
that one or more of these reasons apply does not necessarily mean that
automating moral decisions is a good idea.  But when AI systems do end
up making these decisions, we should not close our eyes to the fact
that they have a moral component, or na\"ively think that we can
always effectively bring humans into the loop to make these decisions.
The best way for AI systems to learn how to make these decisions may
be by observing examples of human moral decision making, but in the
end, they are likely to have to make individual decisions themselves.


\begin{thebibliography}{10}

\bibitem{Abraham07:Clearing}
David Abraham, Avrim Blum, and Tuomas Sandholm.
\newblock Clearing algorithms for barter exchange markets: {E}nabling
  nationwide kidney exchanges.
\newblock In {\em Proceedings of the ACM Conference on Electronic Commerce
  (EC)}, pages 295--304, San Diego, CA, USA, 2007.

\bibitem{Arora11:Global}
Nimar~S. Arora, Stuart Russell, Paul Kidwell, and Erik~B. Sudderth.
\newblock Global seismic monitoring: A {B}ayesian approach.
\newblock In {\em Proceedings of the National Conference on Artificial
  Intelligence (AAAI)}, pages 1533--1536, San Francisco, CA, USA, 2011.

\bibitem{Brown19:Superhuman}
Noam Brown and Tuomas Sandholm.
\newblock Superhuman {AI} for multiplayer poker.
\newblock {\em Science}, 365(6456):885--890, 2019.

\bibitem{Chan20:Artificial}
Lok Chan, Kenzie Doyle, Duncan~C. McElfresh, Vincent Conitzer, John~P.
  Dickerson, Jana~Schaich Borg, and Walter Sinnott-Armstrong.
\newblock Artificial artificial intelligence: Measuring influence of {AI}
  `assessments' on moral decision-making.
\newblock In {\em The Third AAAI/ACM Conference on AI, Ethics, and Society
  (AIES-20)}, pages 214--220, New York, NY, USA, 2020.

\bibitem{Conitzer17:Moral}
Vincent Conitzer, Walter Sinnott-Armstrong, Jana~Schaich Borg, Yuan Deng, and
  Max Kramer.
\newblock Moral decision making frameworks for artificial intelligence.
\newblock In {\em Proceedings of the Thirty-First AAAI Conference on Artificial
  Intelligence}, pages 4831--4835, San Francisco, CA, USA, 2017.

\bibitem{Freedman20:Adapting}
Rachel Freedman, Jana~Schaich Borg, Walter Sinnott-Armstrong, John~P.
  Dickerson, and Vincent Conitzer.
\newblock Adapting a kidney exchange algorithm to align with human values.
\newblock {\em Artificial Intelligence}, 283(103261), 2020.

\bibitem{Hendrycks21:What}
Dan Hendrycks, Mantas Mazeika, Andy Zou, Sahil Patel, Christine Zhu, Jesus
  Navarro, Dawn Song, Bo~Li, and Jacob Steinhardt.
\newblock What would {J}iminy {C}ricket do? {T}owards agents that behave
  morally.
\newblock In {\em NeurIPS Datasets and Benchmarks}, 2021.

\bibitem{Kahng19:Statistical}
Anson Kahng, Min~Kyung Lee, Ritesh Noothigattu, Ariel~D. Procaccia, and
  Christos-Alexandros Psomas.
\newblock Statistical foundations of virtual democracy.
\newblock In {\em Proceedings of the 36th International Conference on Machine
  Learning (ICML-19)}, pages 3173--3182, Long Beach, CA, USA, 2019.

\bibitem{Leibo21:Scalable}
Joel~Z. Leibo, Edgar~A. Du\'e{\~n}ez-Guzm\'an, Alexander Vezhnevets, John~P.
  Agapiou, Peter Sunehag, Raphael Koster, Jayd Matyas, Charlie Beattie, Igor
  Mordatch, and Thore Graepel.
\newblock Scalable evaluation of multi-agent reinforcement learning with
  {M}elting {P}ot.
\newblock In {\em Proceedings of the 38th International Conference on Machine
  Learning (ICML-21)}, pages 6187--6199, 2021.

\bibitem{Letchford08:Ethical}
Joshua Letchford, Vincent Conitzer, and Kamal Jain.
\newblock An ethical game-theoretic solution concept for two-player
  perfect-information games.
\newblock In {\em Proceedings of the Fourth Workshop on Internet and Network
  Economics (WINE)}, pages 696--707, Shanghai, China, 2008.

\bibitem{McElfresh21:Indecision}
Duncan McElfresh, Lok Chan, Kenzie Doyle, Walter Sinnott-Armstrong, Vincent
  Conitzer, Jana~Schaich Borg, and John Dickerson.
\newblock Indecision modeling.
\newblock In {\em Proceedings of the Thirty-Fifth AAAI Conference on Artificial
  Intelligence}, pages 5975--5983, Virtual conference, 2021.

\bibitem{Noothigattu17:Making}
Ritesh Noothigattu, Snehalkumar `Neil'~S. Gaikwad, Edmond Awad, Sohan D'Souza,
  Iyad Rahwan, Pradeep Ravikumar, and Ariel~D. Procaccia.
\newblock A voting-based system for ethical decision making.
\newblock In {\em Proceedings of the Thirty-Second AAAI Conference on
  Artificial Intelligence}, pages 1587--1594, New Orleans, LA, USA, 2018.

\bibitem{Roth04:Kidney}
Alvin~E. Roth, Tayfun Sonmez, and M.~Utku {\"U}nver.
\newblock Kidney exchange.
\newblock {\em Quarterly Journal of Economics}, 119(2):457--488, 2004.

\bibitem{Subercaseaux23:Packing}
Bernardo Subercaseaux and Marijn J.~H. Heule.
\newblock The packing chromatic number of the infinite square grid is 15.
\newblock In {\em TACAS 2023: Tools and Algorithms for the Construction and
  Analysis of Systems}, pages 389--406, 2023.

\end{thebibliography}
\end{document}